\documentclass[pdflatex,sn-mathphys-num]{sn-jnl}% Math and Physical Sciences Numbered Reference Style
\usepackage{graphicx}%
\usepackage{subcaption}
\usepackage{multirow}%
\usepackage{amsmath,amssymb,amsfonts}%
\usepackage{amsthm}%
\usepackage[title]{appendix}%
\usepackage{xcolor}%
\usepackage{textcomp}%
\usepackage{manyfoot}%
\usepackage{booktabs}%
\usepackage{algorithm}%
\usepackage{algorithmicx}%
\usepackage{algpseudocode}%
\usepackage{listings}%
\usepackage{xspace}
\usepackage[export]{adjustbox}
\usepackage{tabularx}
\usepackage{changepage}
\theoremstyle{thmstyleone}%
\theoremstyle{thmstyletwo}%

\theoremstyle{thmstylethree}%

\begin{document}

\title[Article Title]{Fully AI-Generated Image Detection: Definition, Recent Advances and Challenges}
%for Social Trust Generalizable
%from the Perspective of Image Authenticity
% Detecting Images Generated by Deep Generative Models

\author[1,2]{\fnm{Qijie} \sur{Xu}}\email{qijxu@zju.edu.cn}
\author[1,2]{\fnm{Can} \sur{Wang}}\email{wcan@zju.edu.cn}
\author[1,2]{\fnm{Jiawei} \sur{Chen}}\email{sleepyhunt@zju.edu.cn}

\author[3]{\fnm{Siwei} \sur{Lyu}}\email{siweilyu@buffalo.edu}
\author*[3]{\fnm{Defang} \sur{Chen}}\email{defangch@buffalo.edu}

\affil[1]{\orgdiv{State Key Laboratory of Blockchain and Data Security}, \orgname{Zhejiang University}, \orgaddress{%\street{Street}, 
\city{Hangzhou}, %\postcode{100190}, \state{State}, 
\country{China}}}

\affil[2]{\orgname{Hangzhou High-Tech Zone (Binjiang) Institute of Blockchain and Data Security}, \orgaddress{%\street{Street}, 
\city{Hangzhou}, %\postcode{100190}, \state{State}, 
\country{China}}}

\affil[3]{\orgdiv{Institute for Artificial Intelligence and Data Science}, \orgname{University at Buffalo, State University of New York}, \orgaddress{%\street{Street}, 
\city{Buffalo}, %\postcode{610101}, 
%\state{State}, 
\country{USA}}}

% \affil[4]{\orgdiv{Department of Computer Science and Engineering}, \orgname{University at Buffalo, State University of New York}, \orgaddress{%\street{Street}, 
% \city{Buffalo}, %\postcode{610101}, 
% %\state{State}, 
% \country{USA}}}

\abstract{
Recent advances in visual generative models have enabled the creation of highly realistic, fully AI-generated images without relying on real source content. While beneficial for many applications, these models also pose significant societal risks, as they can be easily exploited to produce convincing Deepfakes. Detecting them represents a foundational yet challenging problem in AI media forensics, requiring detectors to reliably extract the inherent artifacts imprinted by generative architectures.
In this Review, we provide a systematic overview of fully AI-generated image detection. Following the standard detector design pipeline, we focus on two key components: dataset construction and artifact extraction. We analyze how dataset design influences the generalization and robustness of learned artifacts, and categorize existing artifact extraction methods based on the primary inductive priors leveraged to isolate artifacts. Within this framework, we systematically review existing works. Finally, we highlight open problems and envision several future directions for developing more robust and generalizable detectors.
Reviewed works in this survey can be found at \url{https://github.com/zju-pi/Awesome-Fully-AI-Generated-Image-Detection}.
}

\maketitle

\section{Introduction}
\label{sec:introduction}

Recent years have witnessed explosive growth in visual generative AI, such as Generative Adversarial Networks (GANs)~\cite{goodfellow2014generative} and diffusion models \cite{ho2020denoising,rombach2022high}.
These advancements have led to remarkable gains in fidelity and diversity, enabling widespread adoption across creative, scientific, and industrial applications \cite{guo2024diffusion,weiss2023guided,lee2025generative}.

\begin{figure*}
    \centering
    \makebox[\textwidth][c]{
        \begin{minipage}{174mm}
            \centering
            \includegraphics[width=174mm, height=234mm, keepaspectratio]{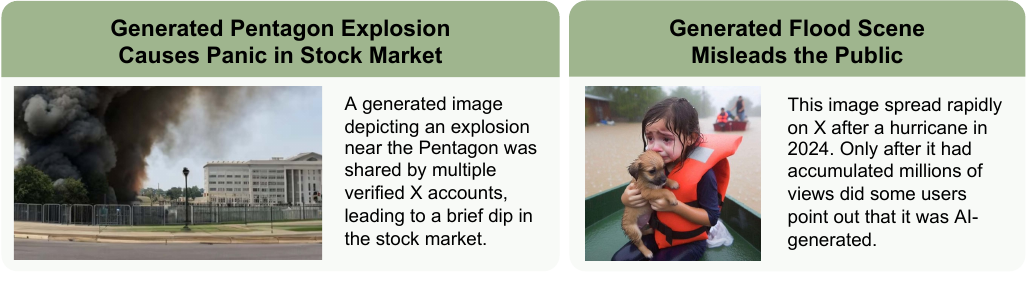}
            \caption{
            Two real-world examples demonstrating the severe societal impact of highly realistic fully AI-generated images, showing that such images can now effortlessly mislead the public and trigger profound real-world consequences.\protect\footnotemark[1]\protect\footnotemark[2]
            }
            \label{fig:harm}
        \end{minipage}   
    }
\end{figure*}

At the same time, the increasing ability to synthesize highly realistic images raises serious societal and ethical concerns. These advanced technologies can be misused to produce deceptive images, facilitating misinformation, blackmail, financial fraud, and other illegal activities~\cite{lyu2020deepfake,farid2025mitigating,amerini2025deepfake}.
Earlier AI-driven image manipulation predominantly relied on partial-editing techniques, such as face swapping or attribute manipulation \cite{pei2024deepfake}. Because these approaches modify specific regions within real source images, they are inherently constrained by the availability and original content of the source material.
However, recent advances in generative models have fundamentally changed this paradigm by enabling the synthesis of high-quality images at the whole-image level. These fully-generated images can be created at scale, depicting arbitrary people, objects, scenes, or completely fabricated ``events'' without requiring a real source image. Achieving an unprecedented level of perceptual realism, they effortlessly deceives the human eye, making them easier to weaponize for large-scale misinformation and much harder to trace or refute.
Such images now circulate widely online and have already led to significant real-world consequences, as shown in Figure \ref{fig:harm} for example. In one notable instance of malicious exploitation, a fabricated image depicting an explosion near the Pentagon was amplified by verified social media accounts, triggering widespread panic and a tangible dip in the stock market. Another instance shows a highly realistic fully AI-generated image of flooding that circulated rapidly after a hurricane. Although it was shared by users who genuinely mistook it for an authentic photograph, this case underscores the profound potential of AI generation not merely to inadvertently mislead the public, but to be weaponized in cognitive warfare, severely eroding institutional credibility and destabilizing societal trust during major events.

\footnotetext[1]{\url{https://edition.cnn.com/2023/05/22/tech/twitter-fake-image-pentagon-explosion/index.html}}
\footnotetext[2]{\url{https://x.com/AmyKremer/status/1841928828576272548}}

These developments have motivated growing interest in detecting fully AI-generated images.
Fundamentally, the task of detecting fully AI-generated images serves as the foundational problem of AI media forensics. Because a broad spectrum of modern media manipulations---spanning from localized edits to fully generated scenes, and from static images to highly realistic videos---are increasingly driven by general-purpose visual generative models, a deep understanding of artifacts produced by these models is beneficial to effectively detecting all these threats. Lacking authentic content, fully AI-generated images provide an environment free from interferences of real images to better investigate these artifacts. Therefore, advancing fully AI-generated image detection will not only address the current whole-image threats but also establish a theoretical groundwork for combating all forms of AI manipulation. Beyond its foundational importance, the proliferation of deceptive fully-generated images makes tackling this problem a pressing practical necessity.
However, accomplishing this task requires a fundamental paradigm shift in detection methodologies. A portion of existing detectors designed for partially edited images relies on manipulation traces, such as inconsistencies between manipulated and untouched regions \cite{pei2024deepfake} and blending artifacts \cite{li2020face,shiohara2022detecting}. However, fully-generated images lack such traces, rendering these methods less effective in practice \cite{tan2024rethinking,zhu2023genimage} and less justified in theory. Consequently, the detection of fully-generated images is forced to rely on the inherent artifacts imprinted by the generative models.
However, extracting such inherent artifacts poses a formidable challenge. Rapidly evolving generative architectures continually invalidate previously identified artifacts; boundless image contents and post-processing operations distort true artifacts and introduce spurious shortcuts for detectors. Therefore, to be practically reliable, these artifacts must remain stable across diverse image contents, generative models, and post-processing pipelines, which makes identifying them particularly challenging.

To address these challenges, a growing body of works has explored various detection methodologies to capture generalizable artifacts and rigorous dataset construction strategies to prevent detectors from exploiting spurious shortcuts. However, the rapid evolution of this field has led to a somewhat fragmented landscape, with differing perspectives on what constitutes effective artifacts and how datasets should be constructed to support the training of detectors.
In this Review, we provide a structured overview of recent progress in fully AI-generated image detection to help researchers effectively navigate the overall landscape. Following the practical pipeline of detector design, we organize the discussion into two components. First, we examine \textbf{dataset construction} in Section \ref{sec:dataset_construction}, emphasizing its critical role in enabling detectors to learn generalizable and robust artifacts. We summarize key principles and common approaches for building effective datasets. Second, we review methods for \textbf{artifact extraction} in Section \ref{sec:detection}, proposing a taxonomy based on the primary inductive priors these approaches leverage to extract generative artifacts. We provide a comprehensive summary of existing artifact extraction approaches within this taxonomy. Finally, we highlight open challenges and outline future directions in Section \ref{sec:future_direction} to develop more generalizable and robust detectors.

\begin{table}[!tbp]
\makebox[\textwidth][l]{
\hspace*{-0.65cm}
\renewcommand{\arraystretch}{1.3}
\begin{tabular}{
>{\centering\arraybackslash}m{3cm}
>{\centering\arraybackslash}m{3cm}
m{6.5cm}
}
\toprule
\multicolumn{2}{c}{\bf Operator Types} & {\bf Examples} \\
\midrule
\multirow{3}{*}[-5ex]{$\mathrm{I}.$ \textbf{Generation}} & Real-world recording & digitizing real-world scenes via camera sensors \\
\cmidrule{2-3}
& Computer Graphics (CG) rendering & rasterizing of 2D or 3D models, cartoon animations from CG, MIDI generated music \\
\cmidrule{2-3}
& AI synthesis & synthesis tools based on variants of GAN models, VAE models, diffusion models, autoregressive models, and diffusion transformer models \\
\hline
\multirow{3}{*}[-3ex]{$\mathrm{II}.$ \textbf{Editing}} & Adding/replacing & restoration (denoising, super-resolution), splicing (non-AI variants or AI-enabled variants including face-swap, lip-syncing, and object/face attribute editing), smart fill, region cloning \\
\cmidrule{2-3}
& Removal & seamless carving, AI-enabled object removal\\
\cmidrule{2-3}
& Stylizing & pixelization, style transfer (\emph{e.g.}\xspace cartoonish, painting-like, or any artistic styles)\\
\hline
\multirow{2}{*}[-7.5ex]{$\mathrm{III}.$ \textbf{Transformation}} & Geometric & rotation, up/downsizing, shear\\
\cmidrule{2-3}
& Signal & blurring (low-pass filtering), vignetting, refocusing, sharpening (high-pass filtering), noise injection, color transform (white-balance adjustment, tone mapping, color highlighting), contrast transform (brightness adjustment, histogram equalization, gamma correction), Bayer mosaic / demosaic
\\
\hline
\multicolumn{2}{c}{$\mathrm{IV}.$ \textbf{Injection}} & adversarial perturbation, watermarking, steganography\\
\hline
\multicolumn{2}{c}{$\mathrm{V}.$ \textbf{Formatting}} & JPEG, PNG, TIFF\\
\bottomrule
\end{tabular}
}
\caption{Types of media operations and examples, including $\mathrm{I}.$ Generation, $\mathrm{II}.$ Editing, $\mathrm{III}.$ Transformation, $\mathrm{IV}.$ Injection, $\mathrm{V}.$ Formatting.}
\label{tab:media_operators_no_type}
\end{table}

\section{Problem Definition}

Before delving into specific dataset construction strategies and artifact extraction methodologies, it is essential to establish a precise scope for the task and identify the core properties that detectors must possess. In Section \ref{subsec:definition_fully}, we formalize the definition of fully AI-generated images by introducing a structural model of the image processing pipeline, explicitly distinguishing them from partial editing. Subsequently, in Section \ref{subsec:definition_objective}, we detail the two fundamental challenges that dictate detector design and evaluation: achieving cross-model generalization against rapidly evolving generative architectures, and maintaining robustness against interferences.

\subsection{Fully AI-Generated Image Detection}
\label{subsec:definition_fully}

To establish a unified formulation for fully AI-generated image detection and systematically conceptualize how post-processing interfaces with this task, we first introduce a general framework that models the image processing pipeline.

An image can be viewed as the terminal outcome of a sequence of processing operations. This process can be described procedurally as follows:

\begin{itemize}
    \item \textbf{Initialization}: The process begins with one or more independent generation operations (see Table \ref{tab:media_operators_no_type} for details and examples), which produce initial digital images.
    \item \textbf{Processing}: Various combinations of non-generation operations is applied to the initial images to produce intermediate results, including editing, transformation, injection, and formatting (see Table \ref{tab:media_operators_no_type}), which are referred to as post-processing operations. Post-processing operations can also be applied to intermediate results. Crucially, editing operators are the unique class of operations that can take multiple images as input. 
    \item \textbf{Output}: The process terminates with a single output image, which may result from any type of operator.
\end{itemize}

This procedural formulation naturally induces a directed acyclic graph (DAG) representation of the image processing pipeline. In this graph, each node corresponds to an operation and its resulting image, and edges represent data dependencies between operations. Nodes with an in-degree of zero correspond to generation operators. Nodes with an in-degree greater than one correspond to editing operations that combine information from different sources. The final output image is represented by the unique node with an out-degree of zero. This topological framework allows us to characterize different image synthesis mechanisms based on their pipeline structures.

\begin{figure}[htbp]
    \centering

    \makebox[\textwidth][c]{
    \begin{minipage}{174mm}
        \centering
        
        \begin{subfigure}{\linewidth}
            \centering
            \includegraphics[width=\linewidth, max height=42mm, keepaspectratio]{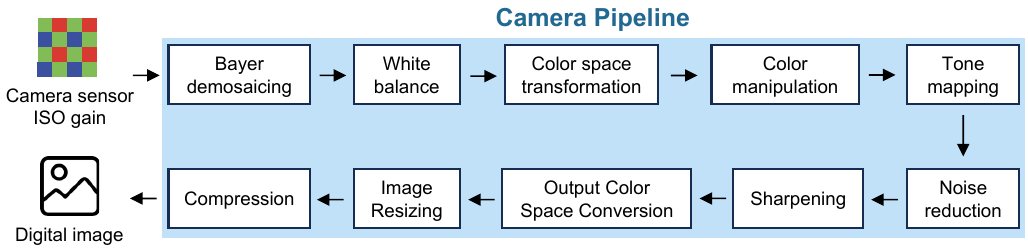}
            \caption{}
        \end{subfigure}
        
        % \vspace{1em}
        
        \begin{subfigure}{0.365\linewidth}
            \centering
            % \raisebox{0.06cm}{\includegraphics[width=\linewidth,max height=40mm, keepaspectratio]{figure/capture_scene.pdf}}
            \includegraphics[width=\linewidth,max height=48mm, keepaspectratio]{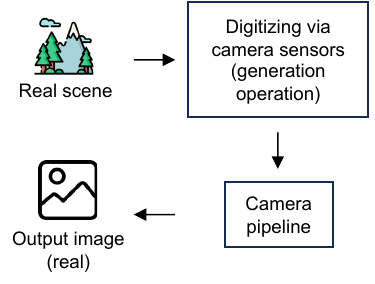}
            \caption{}
        \end{subfigure}
        \hfill
        \begin{subfigure}{0.55\linewidth}
            \centering
            \includegraphics[width=\linewidth, max height=48mm, keepaspectratio]{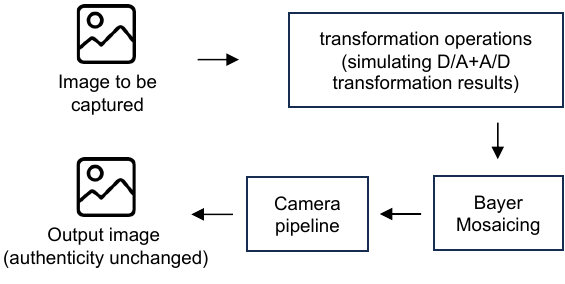}
            \caption{}
        \end{subfigure}
        
        % \vspace{1em}
        
        \begin{subfigure}{\linewidth}
            \centering
            \includegraphics[width=\linewidth, max height=42mm, keepaspectratio]{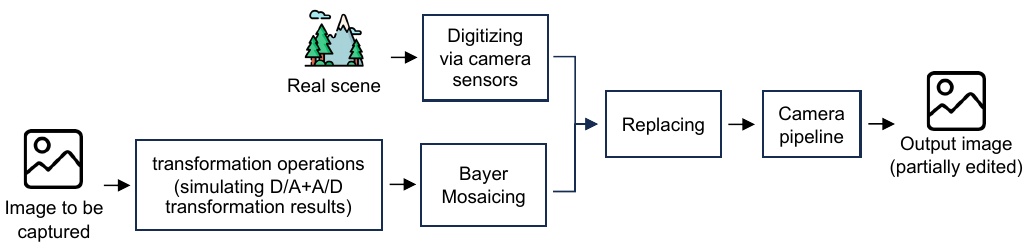}
            \caption{}
        \end{subfigure}

        \caption{Directed Acyclic Graph (DAG) representations of camera capture under different scenarios. (a) \textbf{The camera pipeline} \cite{delbracio2021mobile}. Camera capture is not a singular operation; it comprises a sequence of post-processing transformations from sensor ISO gain to the final image. (b) \textbf{Photographing a real scene}. The process begins with a real-world recording generation node, followed by the camera pipeline. (c) \textbf{Photographing an existing image} (re-digitization \cite{li2025bridging}). Crucially, real-world recording strictly applies only to real-world scenes lacking a prior digital version within the current pipeline. The process of converting an existing digital image to an analog signal (printing/displaying) and recapturing it can be equivalently simulated by a sequence of transformation operations to its digital version in the pipeline, preserving the image's original authenticity. For instance, color/contrast transforms and blurring simulate the printer characteristics, while geometric transformations, vignetting, color/contrast transforms, noise injection, and Bayer mosaicing simulate the lighting, optics and sensor properties. (d) \textbf{Photographing a real scene containing an image}. This can be modeled by initiating a real-world recording for the authentic background and applying the transformations from (c) to the existing image. These two branches merge via a replacing operation before passing through the camera pipeline, rendering the output a partially edited image.}
        \label{fig:camera}
    \end{minipage}
    }
\end{figure}

This Review focuses strictly on the detection of fully AI-generated images, explicitly excluding partially edited images. Within our image processing pipeline framework, we restrict our attention to images whose processing pipeline contains no editing operations. Since only editing operations can take multiple images as input, excluding them strictly reduces the DAG to a single linear chain originating from a single generation node. Within this restricted setting, an image is defined as ``fully AI-generated'' if its singular generation operator corresponds to AI synthesis. The task of fully AI-generated image detection is thus to determine whether the generation operator in such a sequential pipeline corresponds to AI synthesis. Specifically, obtaining an image via camera capture is not inherently a generation operation, and its structural role within the image processing pipeline depends on the captured content; see Figure \ref{fig:camera} for details.

\subsection{Desired Properties of Fully AI-Generated Image Detection}
\label{subsec:definition_objective}

\subsubsection{Generalization Across Different Generative Models}
\label{subsubsec:objective_generalization}

An effective, fully AI-generated image detector should generalize across a broad spectrum of generative models. Ideally, a reliable detector ought to identify images synthesized by any known generative model to maintain practical utility.
However, this requirement exposes a critical vulnerability: as generative models evolve rapidly, detectors risk becoming obsolete and may fail to recognize outputs from emerging architectures. To understand this challenge, a common baseline involves directly adopting standard image classification frameworks (i.e., universal deep learning backbones, such as ResNet \cite{he2016deep} or Vision Transformer \cite{dosovitskiy2020image}, trained with binary cross-entropy loss) to train a binary classifier on real and fully-generated images. By intentionally restricting their training data to a limited set of generative models, this setup simulates the real-world scenario of encountering unknown models. Although such frameworks achieve near-perfect accuracy on images produced by the model seen during training, they generalize poorly across model families: a classifier trained on GAN-generated images often fails to detect diffusion-generated images, and retraining on diffusion models does not restore generalization across GANs or even across different diffusion variants \cite{ojha2023towards,corvi2023detection}.

These failures occur because standard classification frameworks essentially reduce detection to memorizing seen model fingerprints \cite{ojha2023towards}. Consequently, ensuring their generalizability can require continuous data collection and frequent retraining when novel generative models emerge. However, this reactive approach faces prohibitive data collection costs due to the highly fragmented landscape of customized and fine-tuned models \cite{abdullah2024analysis} and restricted closed-source commercial APIs. Moreover, it may leave a critical ``zero-day'' vulnerability window, failing to recognize images from newly emerged generative models until new data is acquired and detectors are retrained.

Therefore, research on fully AI-generated image detection generally seeks to endow detectors with inherent generalization capabilities by capturing model-agnostic artifacts. Admittedly, absolute generalization to all future unknown models is theoretically impossible to guarantee, as future generation architectures may fundamentally eliminate the specific artifacts we currently rely on. Nevertheless, diverse generative model architectures often share commonalities, such as underlying neural network architectures. Therefore, there remains a high probability that broadly generalizable artifacts still exist. Exploiting these shared artifacts provides the strongest possible proactive defense against future model advances, avoiding the prohibitive data collection costs and frequent retraining of the standard image classification framework.

\noindent \textbf{Evaluation protocol of generalization.}
To rigorously assess the inherent generalization capabilities, a standard evaluation protocol requires training detectors on images from a limited set of generative models (often just a single model), and testing them on images synthesized by unseen models \cite{wang2020cnn,corvi2023detection,wang2023dire,ojha2023towards}.
While success under this setting cannot guarantee absolute generalization across tomorrow's unforeseen generative models, it provides crucial empirical evidence. When high cross-model accuracy is achieved using so few training architectures, it suggests that the detector has successfully captured universal generative essence rather than merely memorizing generator-specific fingerprints, implying a greater potential to generalize to unknown models.

\subsubsection{Robustness Against Interference}
\label{subsubsec:objective_robustness}

Generated image detectors are also vulnerable to detection-irrelevant interferences---factors that degrade performance without reflecting the intrinsic authenticity of an image. The most widely studied sources of interference are \textbf{post-processing operations} and \textbf{image content}. These factors undermine detection reliability through two distinct mechanisms.

\medskip
\noindent \textbf{Interfering factors directly alter artifact features.}
Post-processing operations can severely distort image statistics and obscure generalizable artifacts \cite{wang2020cnn,ricker2024aeroblade}.
For example, detection accuracy generally drops when input images undergo operations such as JPEG compression \cite{wang2020cnn,ojha2023towards,yan2025sanity}, resizing \cite{ojha2023towards,chen2024drct,karageorgiou2025any,zhong2025beyond}, blurring \cite{wang2020cnn,ojha2023towards,chu2025fire,li2025improving} and Gaussian noise injection \cite{ricker2024aeroblade,cheng2025co}, with heavier perturbations causing more severe performance degradation, making these routine operations a potent tool that attackers can deliberately exploit to evade detection.
Furthermore, artifacts can be content-dependent. As shown in~\cite{zheng2024breaking}, spectral artifacts observed in the noise residuals of the face-oriented latent diffusion model (LDM) \cite{rombach2022high} may vanish when the same architecture is trained on broader datasets like LAION~\cite{schuhmann2021laion}, indicating that artifacts produced by a specific generative model are not unconditionally stable across different semantic spaces. Moreover, detector performance can vary significantly across content \cite{sha2023fake,zheng2024breaking}, further suggesting that the artifact learned by detectors may be content-dependent.

\medskip
\noindent \textbf{Detectors may learn interference signals instead of artifacts.}
In addition to perturbing true artifacts, interfering factors can introduce signals that are useful for distinguishing real and generated images in the training set, yet are entirely unrelated to image authenticity. Detectors may exploit them as shortcuts.
For example, DIRE \cite{wang2023dire}, initially reported as robust, was later found to heavily rely on a JPEG compression bias: generative models outputted PNG files, whereas real images were JPEG-compressed. This discrepancy was propagated through preprocessed reconstruction errors and learned representations \cite{ricker2024aeroblade}.
Similarly, resolution-oriented operations (e.g., resizing, cropping) can introduce resizing artifacts and expose discrepancies in the original resolutions of real and generated images, leading detectors to overfit to resizing traces or resolution-related signals \cite{grommelt2024fake}. In particular, resolution-oriented operations occur not only within the image-processing pipeline, but also inside the detector itself when detectors convert all inputs to a unified resolution required by the backbone network.
Similar concerns apply to image content. For example, if all real images in the training set depict cars while generated ones depict human faces, the model will learn semantic distinctions rather than image authenticity~\cite{abdullah2024analysis}.

\medskip
These vulnerabilities underscore the critical need for bias-free dataset construction and robust artifact extraction.

\section{Dataset Construction}
\label{sec:dataset_construction}

\begin{figure*}[tbp]
    \centering
    \makebox[\textwidth][c]{
        \begin{minipage}{174mm}
            \centering
            \includegraphics[width=174mm, height=234mm, keepaspectratio]{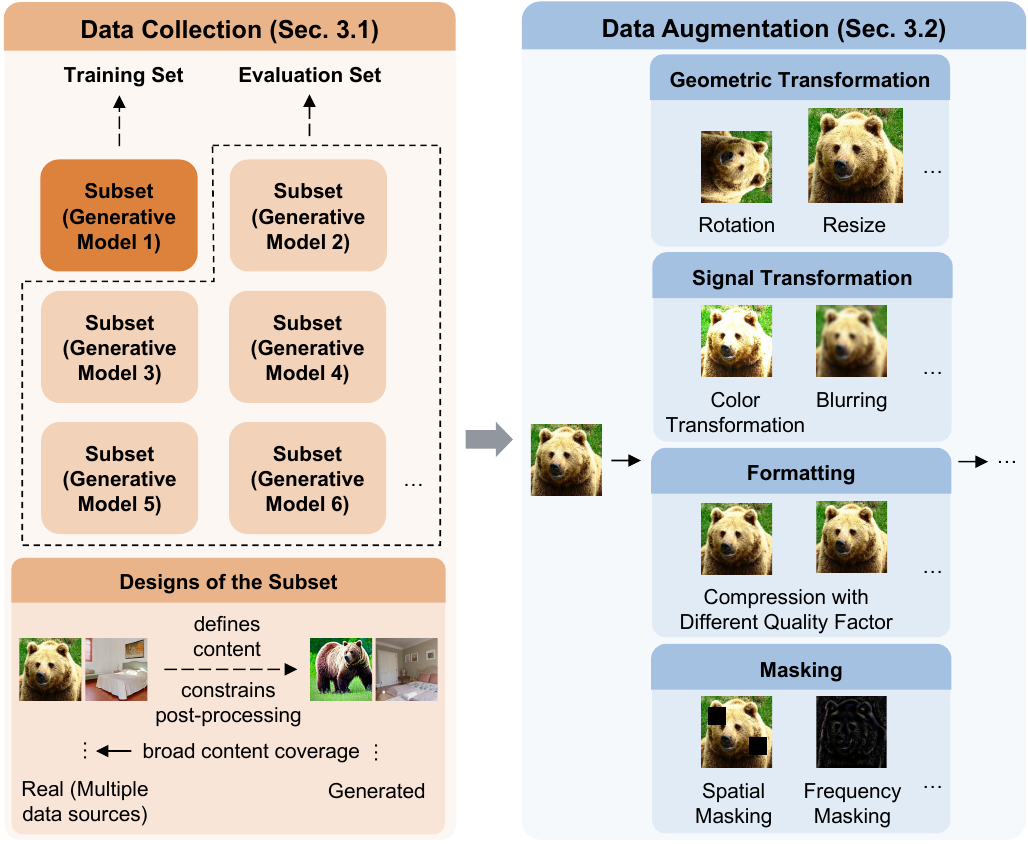}
            \caption{The two-stage dataset construction pipeline for fully AI-generated image detection, illustrating the key designs of each stage.}
            \label{fig:dataset_construction}
        \end{minipage}
    }
\end{figure*}

Dataset design critically shapes the reliability of generated-image detectors. Early studies often underestimated this impact, leading to severe dataset biases and fragile generalization~\cite{wang2023dire,ricker2024aeroblade}, whereas carefully curated datasets can substantially enhance both cross-model generalization~\cite{chen2024drct,guillaro2025bias} and robustness \cite{wang2020cnn,rajan2025effectiveness}. To endow detectors with the desired properties outlined in Section \ref{subsec:definition_objective}, the dataset construction pipeline must systematically fulfill five core objectives: (1) satisfying evaluation protocols for cross-model generalization, (2) content bias mitigation, (3) content coverage, (4) post-processing bias mitigation, and (5) post-processing coverage.
In practice, dataset construction follows a two-stage pipeline: \textbf{data collection} and \textbf{data augmentation}. In this section, we summarize prevailing design choices regarding how these two stages achieve these objectives. An overview of dataset construction is illustrated in Figure \ref{fig:dataset_construction}.

\subsection{Data Collection}
\label{subsec:data_collection}

The data collection stage establishes the fundamental image sets for training and evaluation. This stage typically consists of three procedures: selecting the generative models to be covered by the dataset, collecting real images, and generating images. Below, we present designs of these three procedures that address the aforementioned objectives.

\medskip
\noindent \textbf{Satisfying evaluation protocols.}
Following the evaluation protocol introduced in Section \ref{subsubsec:objective_generalization}, the procedure of selecting generative models typically adopts an asymmetric configuration regarding the number of generative models included. The training set is usually restricted to images synthesized by a single generative model. In principle, there is no strict restriction on which specific generative model serves as the training source, but widely adopted choices include ProGAN \cite{karras2018progressive} (e.g., \cite{wang2020cnn,zhong2023patchcraft,tan2024rethinking,jia2025secret}), ADM \cite{dhariwal2021diffusion} (e.g., \cite{wang2023dire,zhang2024leveraging,chu2025fire}) and Stable Diffusion \cite{rombach2022high} (e.g., \cite{sha2023fake,chen2024drct,guillaro2025bias,zhang2025towards}).
Conversely, the evaluation set must comprehensively reveal a detector's generalizability and expose potential failure modes. Therefore, it should cover a broad range of generative models, spanning open-source and commercial models. Another consideration for broadening model coverage is to alter model hyperparameters such as diffusion sampling steps~\cite{ojha2023towards,tan2024rethinking}. To facilitate fine-grained generalizability analysis, the evaluation set is typically organized into separate subsets for each generative model, containing real images and images generated by that model (see Figure \ref{fig:dataset_construction}).
Finally, to avoid inflated estimates of generalization, some studies \cite{yan2025sanity} manually filter out obvious low-quality synthetic samples after image generation. Without such quality control, distributional discrepancies between easily detectable generated test images and highly realistic real-world outputs can lead to misleading conclusions about the detector's true generalizability~\cite{abdullah2024analysis}.

\medskip
\noindent \textbf{Content bias mitigation.}
To prevent the detector from collapsing into an object classifier, generated images in the training set should be semantically aligned with real ones in the training set. For the evaluation set, maintaining a degree of semantic alignment is also highly beneficial, as it provides a diagnostic environment to ensure the detector evaluates authenticity rather than content biases. This alignment requires a coordinated execution of both collecting real images and generating synthetic ones. Specifically, it involves first selecting appropriate existing real-image datasets (e.g., ImageNet \cite{deng2009imagenet}, COCO \cite{lin2014microsoft}, LSUN \cite{yu2015lsun}, and LAION \cite{schuhmann2021laion}) that match the condition injection mechanisms supported by the chosen generative model, and subsequently synthesizing generated counterparts conditioned on the contents of these real images.
Depending on how the conditioning is applied, existing works fall into distinct generation paradigms to achieve semantic alignment. (1) \textbf{Category-based generation}~\cite{wang2020cnn,wang2023dire,zhang2025towards}. It uses datasets organized by objects or scenes (e.g., LSUN, ImageNet). This paradigm imposes no requirements on the conditioning injection mechanism of the generative model (including unconditional models). (2) \textbf{Text-to-image generation}. It offers finer semantic control by leveraging text-conditioned models to synthesize images from captions found in datasets containing image-text pairs (e.g., COCO, LAION)~\cite{sha2023fake} or generated using captioning models.
While these first two paradigms guarantee concept-level consistency between real and generated images, pushing the semantic alignment further by rendering generated images visually and structurally closer to their real counterparts can help the detector isolate generalizable artifacts from spatial layout biases. This is achieved by (3) \textbf{image-to-image generation}, such as reconstruction \cite{chen2024drct,rajan2025effectiveness,chen2025dual,zhong2025beyond} and inpainting \cite{guillaro2025bias}. Although samples generated via inpainting do not conform to the definition of fully AI-generated images presented in Section \ref{subsec:definition_fully}, they can effectively be utilized as generated training data to aid artifact learning of fully AI-generated images. However, they are generally excluded from the evaluation set.

\medskip
\noindent \textbf{Content coverage.}
Because generated images are semantically aligned with real counterparts, content diversity is directly inherited from the collected real images. Therefore, broad content coverage can be achieved by employing multiple real-image datasets and categories, as well as collecting real images directly from the internet.
Specifically, for the evaluation set, supplementing the aforementioned model-specific, content-aligned data with unconstrained ``in-the-wild'' generated images from social media platforms and AI-creative communities serves as a crucial addition to existing benchmarks \cite{yan2025sanity}. While the strictly organized subsets discussed earlier provide a controlled diagnostic environment, these unconstrained images typically exhibit broader content diversity and higher visual quality, ensuring the overall evaluation set properly approximates the unpredictable distribution of real-world deployment scenarios.

\medskip
\noindent \textbf{Post-processing bias mitigation.}
A critical vulnerability arises from undocumented, built-in post-processing operations inherent in real images collected from existing datasets. These images are fundamentally misaligned with raw AI-generated outputs. While the subsequent data augmentation stage can partially obscure these discrepancies to prevent detectors from leveraging shortcuts (see Section \ref{subsec:data_augmentation}), relying solely on augmentation is insufficient.
For example, applying the same JPEG compression augmentation to both the real and generated images incurs a second round of lossy compression on real images, failing to guarantee that the training set is completely free from compression bias \cite{grommelt2024fake}. Similarly, if resolutions and built-in resizing histories differ between real and generated images, resolution-oriented augmentations may not eliminate the resizing bias. Therefore, some works align the physical properties of generated images with those of their real counterparts and apply targeted post-processing operations to generated images to more closely replicate the post-processing effects in the real counterparts.
For example, the compression alignment problem can be alleviated by estimating the unknown compression quality factors of the real images and applying equivalent lossy compression to generated images \cite{chen2025dual}. For resolution alignment, utilizing image-to-image generation naturally matches resolution between real and reconstructed images, partially alleviating resizing-induced misalignment~\cite{rajan2025effectiveness}. By actively resolving these discrepancies, these approaches yield further improvements in detection performance and robustness compared to relying solely on data augmentation.

\medskip
\noindent \textbf{Post-processing coverage.}
Theoretically, exposing a detector to a wide variety of post-processing operations is achieved in the subsequent data augmentation stage (see Section \ref{subsec:data_augmentation}). However, relying solely on augmentation is insufficient to achieve true post-processing coverage due to the presence of built-in post-processing in collected real images. Because data augmentations are applied on top of existing base post-processing operations, they fail to independently capture the diverse post-processing variants in real-world scenarios \cite{grommelt2024fake}. If a dataset is built upon a single real-image source, it might implicitly restrict the detector to a single, homogeneous base post-processing pipeline. Incorporating more real‑image datasets into both the training and evaluation sets during the collection of real images alleviates this problem, as source-level diversification naturally broadens the foundational coverage of these base operations. Empirical results reveal that when evaluating on real images from different datasets, a detector's performance can fluctuate significantly \cite{cozzolino2024zero}, highlighting the importance of incorporating diverse real-image datasets.

\medskip
Ultimately, the aforementioned strategies for mitigating post-processing biases and expanding post-processing coverage still do not perfectly resolve the fundamental challenges posed by built-in post-processing operations of existing real images. Because the exact processing histories of existing real images remain opaque, retroactive alignments can only approximate true post-processing consistency; as long as real images are post-processed, post-coverage remains anchored to existing post-processing pipelines. A rigorous cure would require collecting entirely new RAW real images---a prohibitively expensive endeavor. Thus, these strategies remain a highly practical and cost-effective compromise.

\subsection{Data Augmentation}
\label{subsec:data_augmentation}

Data augmentation is the final yet critical component of dataset construction. It applies a series of on-the-fly transformations to images during the training process. This mechanism forces the detector to learn robust and generalizable artifacts rather than spurious or non-universal cues at the data level, improving the detector's robustness. Existing data augmentation strategies generally fall into two categories: those targeting post-processing robustness and those explicitly facilitating representation learning.

\medskip
\noindent \textbf{Post-processing coverage and bias mitigation.}
Images collected following the designs in Section \ref{subsec:data_collection} still cover only a narrow range of post-processing operations encountered in real-world deployment. Further incorporating diverse post-processing operations is crucial for improving and systematically assessing the detector’s post-processing robustness \cite{wang2020cnn,li2025improving}. Widely used post-processing augmentations include JPEG compression, Gaussian blur, flipping, rotation, cropping, resizing, Gaussian noise injection, and color jitter (Table \ref{tab:media_operators_no_type} provides a more comprehensive list of post-processing operators).
Beyond expanding coverage, this augmentation also partially mitigates post-processing biases by breaking spurious correlations between authenticity and post-processing within the training set. For example, JPEG compression augmentation destroys the shortcut that compression artifacts are absent in generated images, thereby preventing the detector from classifying all PNG files as generated and all JPEG files as real.
However, this augmentation is not a panacea for improving post-processing robustness. For example, noise-residual artifacts (see Section \ref{subsec:detection_pixel_regularity}) remain highly vulnerable even when JPEG-compressed augmentation is included~\cite{yan2025sanity,zhong2023patchcraft,chen2024single}; sometimes, resizing during training can degrade performance regardless of whether images in the evaluation are resized~\cite{li2025improving}.

\medskip
\noindent \textbf{Facilitating representation learning.}
In addition to simulating real-world interferences, recent work explores augmentation strategies that explicitly promote artifact learning. Masking-based approaches---whether masking spatial regions \cite{jia2025secret,chu2025fire,li2025improving} or specific bands within the frequency spectrum~\cite{doloriel2024frequency}---encourage detectors to utilize information globally across multiple regions. This prevents over-reliance on specific areas, thus improving generalization and robustness.

\begin{figure*}[tbp]
    \centering
    \makebox[\textwidth][c]{
        \begin{minipage}{162mm}
            \centering
            \includegraphics[width=162mm, height=222mm, keepaspectratio]{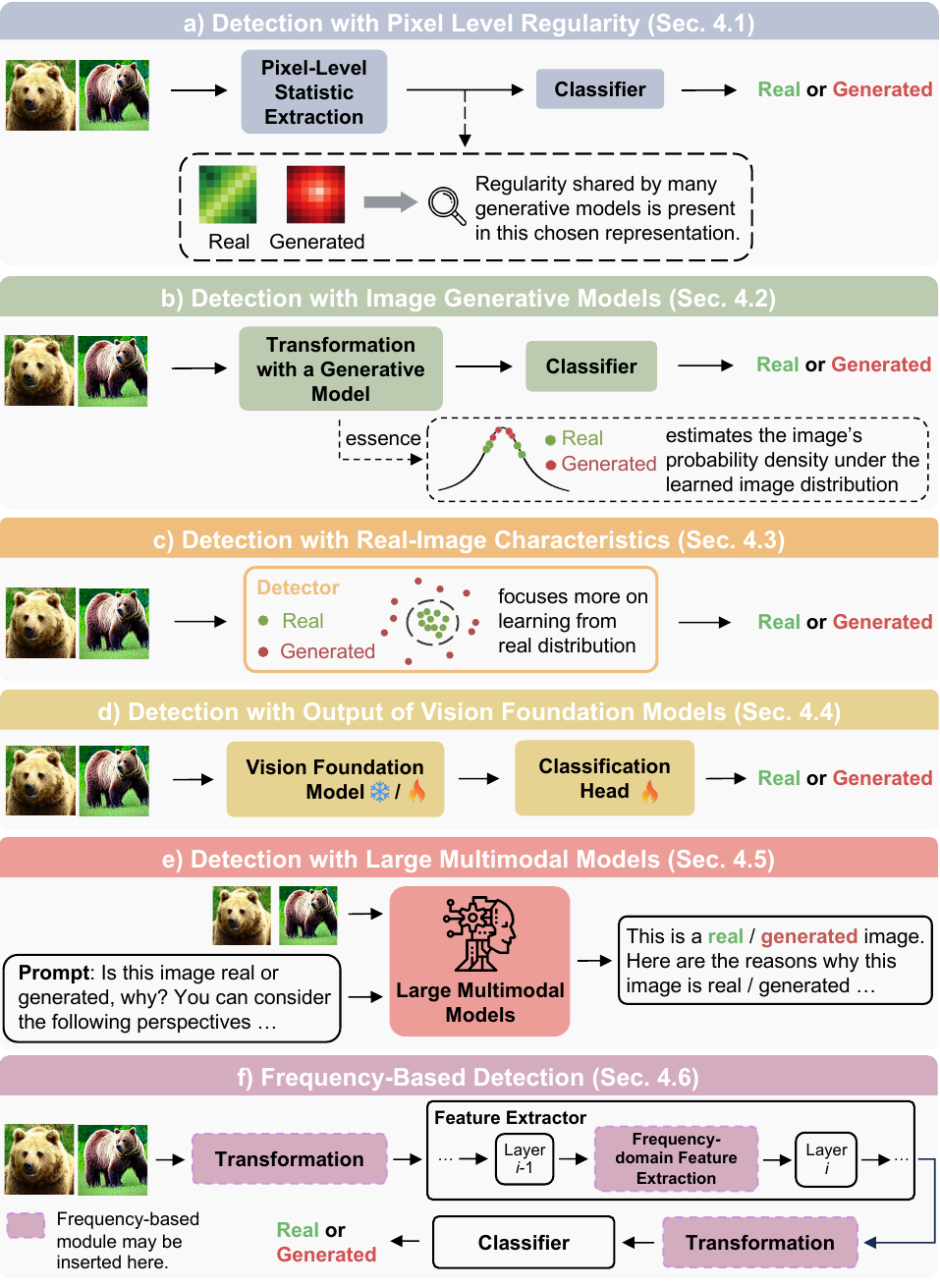}
            \caption{Conceptual overview of six categories of artifact extraction paradigms.}
            \label{fig:artifact_extraction}
        \end{minipage}
    }
\end{figure*}

\section{Artifact Extraction}
\label{sec:detection}

Building upon the constructed dataset, the primary goal of artifact extraction is to learn discriminative representations from a limited set of generative models (often a single one) that successfully generalize to unseen architectures.

To systematically navigate this landscape, we categorize existing artifact extraction approaches based on the primary inductive priors they leverage to isolate artifacts. First, operating primarily within the spatial domain, we organize detection methodologies according to the primary inductive priors used to expose these artifacts: (1) Explicit regularities (Section \ref{subsec:detection_pixel_regularity}). These methods extract statistical anomalies or structural dependencies directly inherent in the pixel values of AI-generated images. (2) Implicit data-distribution priors (Sections \ref{subsec:detection_generative_model} \& \ref{subsec:detection_real}). They quantify deviations between the input image and the underlying distributions of real images or fully AI-generated images. (3) Pretrained generic visual representations (Section \ref{subsec:detection_vision_foundation}). These approaches harness the highly discriminative representations embedded within pretrained vision foundation models. (4) World knowledge and semantic priors (Section \ref{subsec:detection_MLLM}). These methods engage in explicit semantic and physical reasoning by leveraging the rich common-sense prior of Multimodal Large Language Models.
Second, because generative anomalies are often amplified outside the spatial domain, we review frequency-based detection (Section \ref{subsec:detection_frequency}), exploring the frequency domain as a space orthogonal to the spatial domain, where the concepts of the aforementioned methodologies can be adapted to expose deeply concealed artifacts. An overview of these diverse artifact extraction paradigms is illustrated in Figure \ref{fig:artifact_extraction}.

\subsection{Detection with Pixel Level Regularity}
\label{subsec:detection_pixel_regularity}
 
Due to architectural biases and the absence of explicit low-level constraints, generative models may create artificial regularities absent in real images or fail to reproduce subtle patterns intrinsic to real images. Several methods exploit such discrepancies to improve generalization.

One line of work focuses on the base signal of the pixel value. These methods exploit regularities of the base signal at two distinct scales: local dependencies and global statistics. At the local-dependency level, existing methods offer two distinct perspectives on the underlying causes. NPR \cite{tan2024rethinking} and MLEP \cite{yuan2025mlep} attribute local dependencies to an artificial regularity introduced by up-sampling operations in image generation. They capture them through pixel-value differences and distributions within local patches.
Conversely, \citet{liang2025ferretnet} posits that local dependencies reflect the internal consistency of natural textures, and measures deviations between each pixel and the median of its neighbors to quantify this internal consistency. Beyond local dependencies, generative artifacts also manifest in global statistics. For example, real images typically exhibit more uniform color distributions than generated ones~\cite{jia2025secret}.

While the aforementioned methods analyze pixel-level regularities embedded within the base signal, a complementary approach focuses on noise residuals, i.e., pixel-wise fluctuations decoupled from the base signal. Because these subtle noise variations have limited perceptual impact, they are often overlooked during the optimization of generative models, leaving detectable artifacts.
An early approach \cite{liu2022detecting} leveraged CycleISP~\cite{zamir2020cycleisp} to extract noise and identify grid patterns, though these cues generalize poorly to diffusion-generated images \cite{zhong2025beyond}.
Subsequent works \cite{zhong2023patchcraft,yan2025sanity,chen2024single} used SRM-based residuals~\cite{fridrich2012rich}, observing that real images exhibit relatively consistent noise amplitudes across regions, whereas generated images tend to correlate noise strength with texture complexity.
Recent works propose alternative residual constructions, such as treating the last three significant bits of each pixel's binary representation as noise \cite{wang2025lota}, and computing residuals by converting images into other color spaces followed by quantization \cite{fu2025pid}. Although the resulting noise patterns may appear unnatural, they have been empirically shown to be effective.

A critical vulnerability of this class of methods lies in their susceptibility to the interfering factors discussed in Section \ref{subsubsec:objective_robustness}. Specifically, local pixel dependencies and noise residuals can be disrupted by post-processing operations. This inherent fragility underscores the imperative for meticulous data augmentation and comprehensive robustness evaluations when developing detectors based on these signals. Alarmingly, empirical evidence suggests that data augmentation does not universally guarantee robustness. For example, detectors relying on SRM-based noise residuals remain fragile under JPEG compression, even when compression is explicitly included as an augmentation strategy \cite{yan2025sanity,zhong2023patchcraft,chen2024single}.
On the other hand, while global statistical regularities are generally more resilient to perturbations, they may be prone to content dependency, as the statistics of an image can be skewed by its content.

\subsection{Detection with Image Generative Models}
\label{subsec:detection_generative_model}

Generated images are samples from the distribution learned by a generative model. A natural detection strategy is therefore to estimate this learned distribution and assess whether a given image is likely to originate from it.

Diffusion models tend to move samples towards high-probability regions during sampling, suggesting that generated images may lie near local maxima of the learned distribution. However, diffusion models do not directly provide probability densities. Instead, they output the score, i.e., the gradient of the log-density \cite{song2021score}. Leveraging this property, \citet{brokman2025manifold} tests whether an image lies near a local maximum by examining the score magnitude and surface curvature from the score field.

Another line of work evaluates whether a generative model can reproduce the input image. The most common method reconstructs the image using a diffusion model. Several studies \cite{wang2023dire,ricker2024aeroblade,sha2024zerofake,rajan2025effectiveness,cazenavette2024fakeinversion} classify inputs based on the features of reconstruction errors. Rather than directly classifying the error maps, $\mathrm{LARE}^2$ \cite{luo2024lare} leverages the reconstruction error as a spatial attention mechanism. By applying error-derived attention scores to feature maps extracted from the input image, it encourages the detector to focus on poorly reconstructed regions. Beyond reconstruction error, \citet{liang2025denoising} exploits the rate at which the noise-perturbing trajectory diverges from the input image during reconstruction.

A central challenge for this class of methods concerns generalization. Because different generative methods learn different image distributions, detectors built on a specific generative model may struggle to detect images produced by other models. Empirical studies~\cite{liang2025denoising,brokman2025manifold} confirm that some approaches generalize poorly across model classes. For example, detectors relying on Stable Diffusion reconstructions often perform poorly on GAN-generated images. Nevertheless, certain methods still achieve strong cross-model generalization~\cite{liang2025denoising}. Understanding the origins of this behavior remains an open problem. Moreover, a part of methods in this class rely on an iterative multi-step denoising process to perform reconstruction. This introduces extreme inference latency and massive computational overhead, hindering their large-scale deployment on today’s social media platforms, where content is growing explosively.

\subsection{Detection with Real-Image Characteristics}
\label{subsec:detection_real}

As introduced in Section \ref{subsubsec:objective_generalization}, detectors trained under the standard image classification framework often overfit to artifacts specific to generated images. Consequently, they may simply classify images lacking these artifacts as real, which exacerbates generalization issues~\cite{ojha2023towards}. To address this limitation, several studies strengthen the detector’s ability to capture characteristics of real images that define the real-image distribution or manifold. Certain aforementioned approaches can also be interpreted in this way, such as identifying statistical regularities of natural images. In contrast, the methods discussed here aim to learn robust representations of this real-data distribution without relying on explicit prior knowledge of real images.

\subsubsection{Detection with Auxiliary Models Trained Only on Real Data}
\label{subsubsec:detection_OOD}

Models trained exclusively on a specific domain implicitly learn priors over that data distribution. In the context of out-of-distribution (OOD) detection, when such models are exposed to samples outside this distribution, their functional behaviors, such as their effectiveness and robustness in fulfilling intended pretraining objectives, may deviate from those observed on in-distribution (ID) data. Similarly, by treating real images as ID data and fully AI-generated images as OOD samples, evaluating the behavioral deviations of auxiliary models trained exclusively on real images provides discriminative features for fully AI-generated image detection.

\medskip
\noindent \textbf{Reconstruction-based methods.}
Reconstruction-based methods exploit the observation that encoder-decoder frameworks often reconstruct OOD inputs poorly. For example, some works \cite{he2021beyond,cozzolino2024zero} leverage super-resolution models trained solely on real images to recover high-resolution images from down-sampled inputs. These models typically recover real images more faithfully than generated ones.

\medskip
\noindent \textbf{Gradient-based Methods.}
Gradient-based methods examine how model outputs change under small input perturbations. For models trained on ID data, perturbations often induce smaller output changes for ID samples than for OOD samples. RIGID \cite{he2024rigid} observes that, when image embeddings are extracted using the vision foundation model DINOv2 \cite{oquab2024dinov2}, the cosine similarity between a real image and its Gaussian-noise-perturbed version is higher than that for generated images. Subsequent work refines this idea by adopting alternative DINOv2-based scoring mechanisms, such as energy-based estimates of the real-image distribution~\cite{cai2025towards} or self-supervised training losses~\cite{zhang2025detecting}. 
Other studies explore different perturbations, including Haar wavelet decomposition \cite{choi2025training} and transformations used to construct positive samples during DINOv2's unsupervised learning \cite{zhang2025detecting,cai2025towards}.

\medskip
\noindent \textbf{Uncertainty-based Methods.}
Uncertainty-based approaches exploit variability across plausible model parameterizations. While different parameter configurations typically produce consistent outputs on ID samples, they may yield greater variation on OOD samples. 
This phenomenon motivates its application in identifying OOD samples. 
\citet{huang2025diffusion} estimate the posterior distribution of the last-layer parameters in a diffusion model's denoising network using a Bayesian neural network trained on real images. After injecting noise into the input image, the variance of the predicted noise across sampled parameterizations serves as a detection feature.
\citet{nie2025epistemic} report that perturbing pretrained DINOv2 parameters can produce similar uncertainty signals without explicit Bayesian inference.

\medskip
While these auxiliary models provide effective representations of the real-image distribution, a fundamental prerequisite of this paradigm is the availability of powerful auxiliary models trained exclusively on real images. However, general-purpose auxiliary models, such as vision foundation models, do not necessarily exclude AI-generated images during pretraining. As AI-generated images may be incorporated inadvertently into the dataset or even actively included for large-scale pretraining, future auxiliary models may no longer treat generated images strictly as OOD samples. This trend could make it increasingly difficult to adopt more advanced auxiliary models to improve this detection paradigm.

\subsubsection{Real-Centric Representation Learning}

Unlike the methods above, the approaches in this section do not incorporate auxiliary models trained solely on real images. Instead, their training objectives impose stronger constraints on the representations of real images, thereby encouraging the model to encode real-image characteristics when trained jointly on real and generated images.
\citet{zhong2025beyond} design pretext tasks based on real images and their diffusion reconstructions, such as predicting reconstruction errors and noise levels, thereby encouraging the feature extractor to encode deviations from the real-image manifold. A Gaussian mixture model is then employed to model the feature distribution of real images.
\citet{wu2025few} adopts a prototypical network framework that pulls images sharing the same authenticity label closer in feature space, enabling real images to be represented by a prototype.
GenDet \cite{zhu2023gendet} introduces a teacher-student framework in which the discrepancy between teacher and student predictions is minimized for real images and maximized for generated ones.

Unlike approaches in Section \ref{subsubsec:detection_OOD}, which can adopt auxiliary models pretrained on massive datasets, this class of methods models the real-image characteristics directly from the detector's training set. Consequently, the comprehensiveness of the learned real-image characteristics inherently depends on the scale and diversity of the training data, potentially limiting the detector's real-world performance if the training set lacks sufficient variety.

\subsection{Detection with Output of Vision Foundation Models}
\label{subsec:detection_vision_foundation}

As discussed in Section \ref{subsubsec:detection_OOD}, the output embeddings of vision foundation models trained solely on real images (e.g., DINOv2) can exhibit functional behavioral deviations that serve as indicators of the real-image distribution. Beyond evaluating these functional behaviors, a more direct paradigm treats vision foundation models as universal artifact extractors. Remarkably, although these models are trained exclusively on real data and not explicitly optimized for authenticity detection, their output embeddings can directly serve as a highly discriminative feature space for separating real and fully AI-generated images.

\citet{ojha2023towards} were the first to report that applying nearest neighbor and linear probing directly to CLIP-ViT \cite{radford2021learning} image embeddings yields superior generalization than fine-tuning the CLIP-ViT parameters following the standard image classification framework. They attributed this phenomenon to the fact that vision foundation models are not specialized for distinguishing real from generated images; thus, using them without fine-tuning mitigates overfitting to model-specific artifacts. Thereafter, a series of studies leveraged pretrained CLIP-ViT for generated image detection.

Motivated by the fact that CLIP-ViT image embeddings encode high-level semantic content irrelevant to detection, which impedes the extraction of discriminative cues, subsequent studies propose strategies such as using patch-shuffled inputs to suppress semantic content \cite{yang2025d}, removing detection-irrelevant information from image embeddings \cite{zhang2025towards}, and integrating intermediate CLIP-ViT features to emphasize low-level cues \cite{koutlis2024leveraging}.
Furthermore, CLIP-ViT is pretrained to project images into a latent space shared with embeddings of corresponding text descriptions. Several studies observe that incorporating text prompts facilitates adapting image embeddings for generated image detection.
A simple strategy is to follow established methods for transferring CLIP to downstream tasks, which represent real and generated categories by text prompts and classify images based on the cosine similarity between image embeddings and text embeddings \cite{khan2024clipping,liu2024forgery}. Techniques such as learnable prompt contexts \cite{zhou2022learning}, image-conditioned context~\cite{zhou2022conditional}, and further alignment between image embeddings and category-prompt embeddings \cite{liu2024forgery} markedly improve the detector's generalizability \cite{khan2024clipping,liu2024forgery}.
DE-FAKE \cite{sha2023fake} reveals that text-to-image generation tends to generate images strictly adhering to user-provided prompts, whereas real images carry richer details beyond textual descriptions. Therefore, DE-FAKE utilizes the description attached to the image or generated by BLIP \cite{li2022blip}, and extracts artifacts from the concatenation of image embeddings and text embeddings obtained from CLIP-ViT.

However, the conclusion presented in \cite{ojha2023towards} that fine-tuning underperforms adapting embeddings extracted by pretrained parameters does not hold universally. Adopting other fine-tuning strategies, improving the training set as described in Section \ref{subsec:data_collection}, or simply employing vision foundation models other than CLIP-ViT could enable fine-tuning to provide detectors with image embeddings better aligned to generated image detection, while mitigating overfitting induced by fine-tuning.
\citet{guillaro2025bias} show that, when the training set is collected via inpainting, linear probing on embeddings extracted from DINOv2 with registers performs worse than fine-tuning. C2P-CLIP \cite{tan2025c2p} incorporates the authenticity label into each image's text description and fine-tunes CLIP-ViT using the original CLIP contrastive loss. \citet{liu2024mixture} adopts a Mixture-of-Experts (MoE) architecture with a load-balancing loss to encourage diverse feature learning during fine-tuning. Some studies \cite{liu2024forgery,chen2025forgelens,fang2025forensic} insert trainable layers between the modules of vision foundation models while keeping the pretrained parameters frozen to perform fine-tuning.

Despite their strong empirical performance, this class of methods suffers from severe mechanistic opacity in their detection principles. First, it remains fundamentally unclear how these pretrained embedding spaces---which are designed to discard low-level variations in favor of high-level semantics, sometimes explicitly aligning with text embeddings---effectively capture artifacts. Although strategies that suppress semantics improve generalization by forcing the model to focus on low-level traces, this does not fully explain the discriminative power of unmodified embeddings. When a simple classifier is directly applied to these intact high-level representations, it is difficult to determine whether its success stems primarily from implicitly retained low-level artifacts, distinct high-level artifacts (e.g., subtle yet model-detectable semantic discrepancies between certain objects depicted by generative models and their real-world counterparts), or simply exploits hidden content biases.
The theoretical ambiguity also extends to model adaptation strategies, rendering it unpredictable whether freezing or fine-tuning yields better performance. In particular, the underlying reasons why fine-tuning pretrained parameters can sometimes avoid the expected collapse in cross-model generalization remain not fully understood.
Moreover, these opaque mechanisms may be further complicated by potential AI-generated image contamination in the pretraining dataset of the future vision foundation model, similar to the challenges noted in Section \ref{subsubsec:detection_OOD}. Such contamination may cause the vision foundation model to explicitly pull the embeddings of semantically similar real and generated images closer, or to treat certain low-level artifacts as perturbations that should be discarded, thereby further undermining the detector's discriminative capability and confounding its theoretical foundations.

\subsection{Detection with Multimodal Large Language Models}
\label{subsec:detection_MLLM}

Recent advances in multimodal large language models (MLLMs), especially image comprehension capabilities, have enabled their exploitation for generated image detection \cite{huang2025sida,zhou2025aigi,wen2025spot}.
Existing approaches to detecting generated images with MLLMs typically follow the visual question answering framework. The input to the MLLM comprises an image and a prompt composed of a series of visual understanding questions that require the model to assess whether potential artifacts are present in the image from multiple perspectives, including adherence to physical laws, commonsense reasoning, object texture, edge consistency, shifts in color distribution, and so on. The selection of questions may vary depending on the content of images \cite{wen2025spot}.
After encoding images and text prompts into embeddings, the large-language-model-centered reasoning engine performs cross-model inference and reports whether the artifacts described in the prompt are present in the input image, in which region they are present, and the final classification decision via natural language and additional special tokens. AIGI-Holmes \cite{zhou2025aigi} also explores the incorporation of representations extracted by two additional detection methods as part of the input to the reasoning engine.
To enable pretrained MLLMs to produce accurate classification results and generate high-quality responses to the questions, model fine-tuning is necessary. During fine-tuning, one can employ not only binary cross-entropy loss for classification \cite{huang2025sida} but also supervised fine-tuning \cite{huang2025sida,zhou2025aigi,wen2025spot} and preference-optimization \cite{zhou2025aigi} losses used in the post-training of large language models, in which textual annotations required for fine-tuning are primarily obtained by refining automatically generated annotations from pretrained MLLMs.

Unlike the aforementioned methods that primarily rely on statistical or visual feature extraction, detection with MLLMs represents a distinct paradigm driven by semantic reasoning and world knowledge. Through carefully designed prompts, MLLMs can dynamically integrate perspectives ranging from low-level texture anomalies to high-level semantic inconsistencies. This inherent flexibility grants MLLMs remarkable dynamic extensibility, allowing detectors to adapt to emerging forgery traces simply by updating the textual instructions.
Crucially, this paradigm uniquely overcomes the intractable bottlenecks of traditional physical artifact modeling. While a few explicit attempts have been made to model specific physical-law violations, such as geometric inconsistencies \cite{sarkar2024shadows}, these handcrafted rules are notoriously fragile. Explicitly extracting complex physical interactions is algorithmically challenging, and these rigid rules often fail when applied to diverse, unconstrained image contents. In contrast, MLLMs effortlessly bypass these bottlenecks. By leveraging their vast reservoir of prior knowledge and strong common-sense reasoning, they naturally identify violations of complex physical interactions across arbitrary semantic domains.
Furthermore, by articulating these physical anomalies in natural language, MLLMs can provide definitive, ``smoking-gun'' explanations that are highly convincing for human reviewers, offering interpretability unattainable by black-box feature embeddings.
However, detection with MLLMs is not without its limitations. When explaining non-physical artifacts, such as ambiguous texture, unnaturalness, or subtle color shifts, MLLMs often generate vacuous or far-fetched responses that cannot convince users. Driven by their conversational training, they may suffer from hallucination, confidently inventing non-existent flaws to satisfy the prompt rather than providing reliable forensic evidence. Additionally, detection with MLLMs incurs prohibitive computational costs and high inference latency, restricting their deployment in real-time, large-scale content moderation on social media.

\subsection{Frequency-Based Detection}
\label{subsec:detection_frequency}

The training objectives of generative models often neglect frequency-domain information, which causes the discrepancies between real and generated images to be more pronounced in the frequency domain.

Frequency-domain detection methods conceptually mirror the pixel-domain approaches discussed earlier.
Similar to the pixel-domain regularities introduced in Section \ref{subsec:detection_pixel_regularity}, generative models also imprint patterns that are absent from real images in the frequency domain. The most representative pattern is an elevated magnitude of high-frequency components in the generated images. Early studies \cite{frank2020leveraging,durall2020watch} attribute this phenomenon to the up-sampling operations commonly used in generative model architectures. To leverage this regularity, several studies \cite{tan2024frequency,zhang2024leveraging,li2025improving} extract the high-frequency components of images to train the detector. \citet{tan2024frequency} improves the detector's backbone architecture to extract features from both magnitude and phase, achieving superior performance on GAN-generated images. \citet{zhang2024leveraging} highlights the importance of the mask when extracting high-frequency artifacts.
Additionally, some works develop models that treat the frequency spectrum of real images as in-distribution and those of generated images as out-of-distribution, as discussed in Section \ref{subsubsec:detection_OOD}. \citet{karageorgiou2025any} train a spectral reconstruction model with an encoder-decoder architecture on real images. The resulting encoder exhibits markedly different responses to real and generated images.

Furthermore, combining the frequency-domain information with pixel-domain artifacts can further enhance the generalization performance \cite{liu2024forgery,liu2022detecting,chu2025fire}.
\citet{liu2024forgery} incorporates trainable modules between adjacent transformer blocks of CLIP-ViT to incorporate frequency-domain information into image embeddings extracted with the pretrained CLIP-ViT. FIRE \cite{chu2025fire} further improves the generalization capability achieved using diffusion reconstruction error by comparing the reconstruction errors obtained before and after removing the mid-frequency components from the input image.

\section{Open Challenges and Future Directions}
\label{sec:future_direction}

\noindent \textbf{Building a RAW image dataset.}
As introduced in Section \ref{subsec:data_collection}, current alignment strategies can only partially mitigate the hidden biases introduced by the undocumented post-processing of existing real images. A fundamental solution requires completely eliminating these built-in operations.
We therefore advocate curating datasets whose real subsets consist exclusively of RAW images \cite{delbracio2021mobile}, with broad coverage of content and resolution. When using such datasets for training and evaluation, the camera image-processing pipeline~\cite{delbracio2021mobile} and the re-digitization process described in Figure \ref{fig:camera} can be incorporated into data augmentation to simulate realistic camera post-processing and improve robustness.

\medskip
\noindent \textbf{Combining multiple artifact representations.}
As discussed in Section \ref{subsubsec:objective_generalization}, capturing universal generative artifacts is the ultimate objective for ensuring robust detection. However, there is no fundamental constraint dictating that such universal artifacts must manifest within a single feature space. While the methods reviewed in Section \ref{sec:detection} have proposed numerous effective feature spaces, each of these single-view representations inherently captures only a part of the generative traces. The intrinsic differences differentiating real and AI-generated images are highly complex and multi-faceted. Relying exclusively on a single representation leaves detectors unnecessarily vulnerable to emerging architectures that might erase that specific trace.
Therefore, establishing a more robust and proactive defense requires fusing these complementary artifact representations to construct a comprehensive, multi-view characterization of AI synthesis. Existing strategies include concatenating multiple feature representations for the final classification \cite{yan2025sanity}, veto-based fusion~\cite{abdullah2024analysis}, and weighted averaging of spatially aligned representations \cite{corvi2023detection,cheng2025co,fang2025forensic} or logits \cite{zhou2025aigi}. Forensic-MoE \cite{fang2025forensic} further improves integration by fine-tuning the feature extractors used for each representation. Furthermore, as highlighted in Section \ref{subsec:detection_MLLM}, Multimodal Large Language Models (MLLMs) represent a highly advanced paradigm for this integration, leveraging natural language reasoning to dynamically combine these multi-perspective artifacts.

\medskip
\noindent \textbf{Robustness to adversarial attack.}
As deep learning-based classifiers, generated image detectors are inherently vulnerable to adversarial manipulation. While existing studies~\cite{zhou2024stealthdiffusion,abdullah2024analysis,zhou2025breaking} have demonstrated high attack success rates by injecting perturbations either into the generated images or directly into the generative process (e.g., altering initial noise or model parameters), empirical evaluation of adversarial robustness for state-of-the-art detectors remains limited. It is imperative to systematically evaluate the fragility of newly proposed artifact extraction methods under adversarial threats, and to develop proactive defense mechanisms to ensure reliable deployment in adversarial environments.

\medskip
\noindent \textbf{Refining the benchmark.}
Several aspects of current benchmarks warrant further improvement.
(1) Benchmarks should evaluate multi-model training configurations. As discussed in Section \ref{subsubsec:objective_generalization}, the single-model training protocol provides a rigorous proxy to evaluate a detector’s inherent zero-shot generalization. However, practical deployment naturally grants defenders access to training data from multiple known generative models. Leveraging diverse architectures during training can help detectors further marginalize out model-specific fingerprints and learn a more robust, unified decision boundary \cite{park2025community,yang2025d}. Therefore, comprehensive benchmarks should systematically include multi-model training tracks to measure how effectively different artifact representations scale and benefit from diverse training data, ensuring that evaluations mirror real-world detection scenarios.
(2) Performance should be reported separately for real and generated images. Because current benchmarks are predominantly hyper-focused on evaluating generalization across diverse generative models, they typically report only an aggregated overall accuracy for each generator-specific subset of the evaluation set. This practice critically obscures underlying error patterns. To transparently expose a detector's true failure modes, overall accuracy must be split into separate metrics for real and generated subsets, revealing whether a detector systematically misclassifies authentic content (False Positives) or fails to detect outputs from novel generative architectures (False Negatives).
(3) Benchmarks should evaluate sequential post-processing stress tests. While single post-processing operations are routinely evaluated, real-world malicious actors frequently employ complex, sequential chains of content-preserving post-processing operations to launder AI-generated images. Benchmarks must therefore incorporate predefined multi-step post-processing pipelines to evaluate the compounding degradation of the detector's robustness.

\medskip
\noindent \textbf{Shifting from binary classification to forensic reasoning.}
Existing detection frameworks predominantly formulate the task as supervised binary classification, prioritizing simple label assignment over rigorous evidentiary inference. However, in high-stakes, real-world deployments, such as evaluating legal evidence, combating political misinformation, or verifying journalistic integrity, a mere ``fake'' probability score from a black-box model is insufficient to support actionable decision-making. Approaching detection from a forensic perspective is essential, as establishing public and institutional trust requires a verifiable chain of evidence demonstrating precisely why an image is synthetic. Future research could therefore transition from simply assigning discrete labels toward developing interpretable artifact representations and constructing causal explanations for a detector's decisions. For instance, leveraging Multimodal Large Language Models (MLLMs) to identify and articulate violations of physical laws in natural language represents a paradigm that can provide forensic evidence.

\medskip
\noindent \textbf{Extending detection methods to other forgery types.}
Fully AI-generated images represent only one manifestation of modern generative technologies. As established in Section \ref{sec:introduction}, detecting fully AI-generated images serves as the foundational problem of AI media forensics by isolating the intrinsic artifacts of AI synthesis in a pure, interference-free environment. With general-purpose generative models increasingly driving sophisticated partial editing and highly realistic video synthesis, the model-agnostic and robust artifacts discussed in this Review hold immense potential for broader applicability.

\section{Discussion and Conclusion}

While discovering generalizable artifacts is the current imperative, the community will ultimately confront a profound theoretical paradox intrinsic to the passive detection paradigm. Consider the scenario of diffusion reconstruction, in which a real image can, with high probability, be reconstructed with near-zero semantic loss. This presents an insurmountable classification dilemma. If a detector correctly flags the reconstructed image as ``generated'', it inadvertently weaponizes the detector to discredit objective reality and exacerbate the ``Liar's Dividend'', as malicious actors could reconstruct real images to intentionally imprint generative artifacts. Conversely, classifying it as ``real'' concedes that generative models can synthesize technically authentic images and blurs the fundamental boundary between the real and generated manifolds, given that the reconstructed image is fundamentally a sample obtained by starting from noise and proceeding through the diffusion model’s sampling process.

This paradox reveals a fundamental discrepancy between the ultimate goal of media forensics and the current technical paradigm. At its core, authenticity is an epistemological question:
\begin{center}
    \textit{Did the depicted scene actually occur in the physical world?}
\end{center}
Because computer vision alone cannot verify historical events, the field has naturally turned to a proxy task: searching for generative artifacts. While this proxy appears to be a robust substitute, the reconstruction thought experiment exposes that for the vast majority of artifacts, their presence is becoming independent of an image's historical authenticity. If a detector successfully relies on a specific artifact to flag the reconstructed image as ``generated'', it inherently proves that this artifact can be injected into real images without altering their reality. Conversely, if a detector fails to find its targeted artifact in the reconstructed image, it proves that modern generative models are already capable of generating an image that is devoid of this artifact. (Although artifacts constituting physical violations form a notable exception, as they directly refute a scene's historical occurrence and cannot be injected without destroying semantic reality. However, physical plausibility is merely a necessary condition for truth, and generative models are nevertheless increasingly capable of synthesizing scenes devoid of physical flaws.) In either mutually exclusive outcome, the capability boundary of the detector is clear.

Despite this theoretical capability boundary, the methodologies and artifact extraction strategies reviewed in this article remain indispensable. In real-world scenarios, achieving perfect synthesis or seamlessly contaminating real images demands prohibitive computational costs and extreme optimization. The paramount significance of advancing fully AI-generated image detection lies not in achieving absolute invulnerability against a theoretically omnipotent adversary, but in drastically raising the economic and technical thresholds for malicious actors. By continuously uncovering universal, multi-dimensional artifact representations and rigorously hardening them against complex, real-world image distributions and post-processing interferences, the research community is constructing a comprehensive, generalizable, and robust defensive ecosystem. This ecosystem forces adversaries to expend vast resources to evade detection, effectively constituting the foundational first line of defense to stem the scalable proliferation of deceptive media.

\bibliography{sn-bibliography}% common bib file

\end{document}